\pdfoutput=1

\documentclass[11pt]{article}

\usepackage{ACL2023}
\usepackage{times}
\usepackage{latexsym}

\usepackage[T1]{fontenc}

\usepackage[utf8]{inputenc}

\usepackage{microtype}

\usepackage{inconsolata}

\usepackage{multirow}
\usepackage{graphicx}
\usepackage
{bbm}
\usepackage{amsfonts,amssymb}
\usepackage{amsmath}
\usepackage{color}

\title{APOLLO: An Optimized Training Approach for Long-form Numerical Reasoning}

\author{
Jiashuo Sun\textsuperscript{\rm 1},\ \ 
Hang Zhang\textsuperscript{\rm 2},\ \
Chen Lin\textsuperscript{\rm 1}\footnotemark[1],\ \ 
\textbf{
Xiangdong Su\textsuperscript{\rm 3}
Yeyun Gong\textsuperscript{\rm 4},
Jian Guo\textsuperscript{\rm 2}}
\\
\textsuperscript{\rm 1} 
School of Informatics, Xiamen University, China\\
\textsuperscript{\rm 2}
IDEA Research, China\\
\textsuperscript{\rm 3}
College of Computer Science, Inner Mongolia University
\textsuperscript{\rm 4}
Microsoft Research Asia
}
  
\begin{document}
\maketitle
\renewcommand{\thefootnote}{\fnsymbol{footnote}}
 \footnotetext[1]{Corresponding author, chenlin@xmu.edu.cn}
 
\begin{abstract}
Long-form numerical reasoning aims to generate a reasoning program to calculate the answer for a given question. Previous work followed a retriever-generator framework, where the retriever selects key facts from a long-form document, and the generator generates a reasoning program based on the retrieved facts. However, they treated all facts equally without considering the different contributions of facts with and without numerical information. Furthermore, they ignored program consistency, leading to the wrong punishment of programs that differed from the ground truth. In order to address these issues, we proposed APOLLO (\textbf{A}n optimized training a\textbf{P}proach f\textbf{O}r \textbf{L}ong-form numerica\textbf{L} reas\textbf{O}ning), to improve long-form numerical reasoning. APOLLO includes a number-aware negative sampling strategy for the retriever to discriminate key numerical facts, and a consistency-based reinforcement learning with target program augmentation for the generator to ultimately increase the execution accuracy. 
Experimental results on the FinQA~\footnote{\url{https://codalab.lisn.upsaclay.fr/competitions/4138\#results}} and ConvFinQA~\footnote{\url{https://codalab.lisn.upsaclay.fr/competitions/8582\#results}} leaderboards verify the effectiveness of our proposed methods, achieving the new state-of-the-art.~\footnote{Our code is publicly available at \url{https://github.com/GasolSun36/APOLLO}.}

\end{abstract}

\begin{figure}[t]
    \centerline{\includegraphics[width=8cm,keepaspectratio]{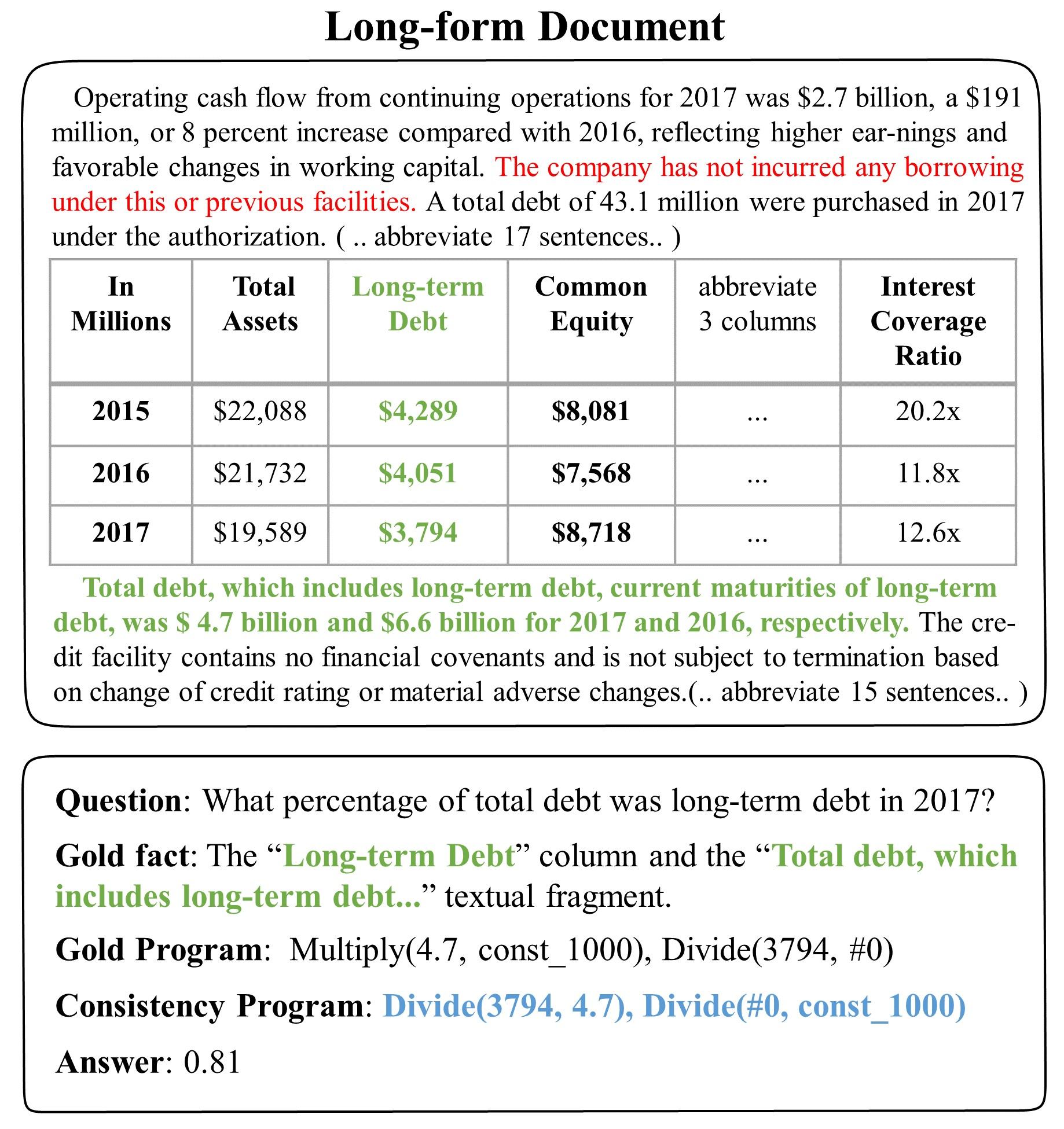}}
    \caption{An example of Long-form Numerical Reasoning. The parameters in gold program are directly from the numerical fact (e.g., \textcolor[RGB]{112,173,71}{table column 2 and the textual fragment in green}) instead of the non-numerical fact (e.g.,  \textcolor[RGB]{255,0,0}{the textual fragment in red}). The answer can be equally generated from the gold program and the \textcolor[RGB]{91,155,213}{consistent program}. \texttt{Const\_x}, \texttt{\#i} denotes constant $x$ and the result of the previous $i-1^{th}$ operator.}
    \label{figure:example}
\end{figure}

\section{Introduction}
Long-form numerical reasoning aims to generate an executable program that answers a specific question. Unlike conventional numerical reasoning tasks \cite{mackenzie2008engine}, long-form numerical operates on a long document (e.g., max $2,679$ tokens in FinQA \cite{chen2021finqa}). Recently, various benchmarks, such as FinQA~\cite{chen2021finqa} and ConvFinQA~\cite{convfinqa}, have been proposed to assess the ability of systems to perform long-form numerical reasoning (Figure~\ref{figure:example}).

A typical framework to solve this task is the retriever-generator question-answering framework, which is firstly introduced by~\citet{chen2021finqa}. 
This framework consists of two stages: training a retriever to identify relevant facts (i.e., a textual fragment or a table column) from the documents and training a generator to generate the executable programs with the retrieved facts. Recently, the use of pre-trained masked language models and ensemble techniques \cite{wang2022novel, zhang2022robustly, wang2022numerical} has led to high accuracy in this task. \citet{zhang2022elastic} and \citet{tablet5} have designed a novel structure for better handling long-form data. \citet{dyrren} has proposed a dynamic retriever-reranker-generator framework to enhance each generation step by a dynamic reranking of retrieved facts. \citet{chatgpt_few_shot} uses large language models (LLMs) such as gpt-3.5-turbo \cite{chatgpt} and GPT-4 \cite{gpt4} as backbone model and zero-shot to inference.

However, existing studies ignore two critical aspects of long-form numerical reasoning. Firstly, they neglect the importance of numerical facts and treat all facts equally. Numerical facts are more important since they are the direct source of parameters in the generated program (as demonstrated in Figure~\ref{figure:example}). Secondly, they disregard the issue of program consistency, where programs can have different expressions but produce the same results. As shown in Figure~\ref{figure:example}, if we use only the gold program as the ground truth in supervised training, the model will wrongly penalize consistent programs.

In this paper, we improve the performance of the retriever-generator framework for long-form numerical reasoning.  We propose a number-aware negative sampling approach for retriever training to prioritize and differentiate between numerical facts (Section~\ref{sec:retriever}). At the generator level, we introduce target program augmentation and consistency-based reinforcement learning to explore the space of consistent programs and improve execution accuracy (Section~\ref{sec:generator}).

The contributions of this work are as follows:

\begin{itemize}
\item We introduce a novel number-aware negative sampling, demonstrating the effectiveness during retriever training for long-form numerical reasoning tasks.
\item We propose consistency-based reinforcement learning and target program augmentation in our generator training to increase execution accuracy.
\item Our approaches achieve the current state-of-the-art of \textbf{72.47} execution accuracy and \textbf{68.01} program accuracy on FinQA, and \textbf{78.76} execution accuracy and \textbf{77.19} program accuracy on ConvFinQA, respectively.
\end{itemize}

\section{Related Work}
\subsection{Retrieval Augmented Generation} 
A series of previous works explore a retrieve-augmented paradigm for text generation \cite{mathqa,wei2022chain,mawps,wang2022novel,zhang2021adversarial,zhang2021poolingformer}.
Popular datasets include NQ \cite{nq}, TriviaQA \cite{triviaqa}, FEVER \cite{fever}, MS-MARCO \cite{ms} and the benchmark KILT \cite{kilt}. The line of this paradigm is, at first, the retriever retrieved amount of passages based on the query~\cite{zhang2023noisy}. Secondly, the generator takes the retrieved passages as input and generates an answer. It is noteworthy that the retrieval-augmented generation approach focuses on retrieving passages from a vast repository of knowledge (typically Wikipedia) as external information due to the limited input provided (mostly only one query). For long-form numerical reasoning tasks, the model needs to focus on retrieving the most critical facts directly from input.

\subsection{Numerical Reasoning} 
\paragraph{MWP Numerical Reasoning}
MWP (Math Word Problem) numerical reasoning is a challenging task that has been introduced for years \cite{liu2020reasoning, sun2023enhancing, huang2024joint}. The task calculates the answer to a short textual question by generating an arithmetic expression. There are several benchmark datasets, including MathQA \cite{mathqa} and MaWPS \cite{mawps}, which all focus on generating target programs for math or calculation problems. The questions in MWP are described in a controlled manner, exhibiting strong regularity. Therefore, some previous works use template-based \cite{wang2019template} or tree-based \cite{jie2022learning, li2023trea} methods. However, the MWP numerical reasoning tasks are mostly general and simple, which only contain one short query and do not need external knowledge and are far different from the tasks we explore in this paper.
\paragraph{Long-form Numerical Reasoning}
Long-form numerical reasoning is more challenging than MWP. The task is to generate the program for a specific question based on retrieved facts from a long-form document. \citet{chen2021finqa} and \citet{convfinqa} have introduced FinQA and ConvFinQA, which are complex question-answering and conversational numerical reasoning tasks for financial reports, respectively. Since long-from numerical reasoning is based on retrieval and generation, therefore some works improve them separately. \citet{wang2022novel} has used DeBERTa \cite{he2021debertav3} to pre-training on financial data. \citet{zhang2022robustly} has proposed an ensemble approach by developing models with different specialized capabilities and fusing their strengths. \citet{wang2022numerical} has devised a cell retriever module to retrieve gold cells to avoid bringing unrelated cells to the generator. \citet{zhang2022elastic} has utilized an adaptive symbolic compiler to generate programs. 
However, previous works overlook the significance of numerical facts and program consistency.

\begin{figure}[t]
    \centerline{\includegraphics[width=8cm,keepaspectratio]{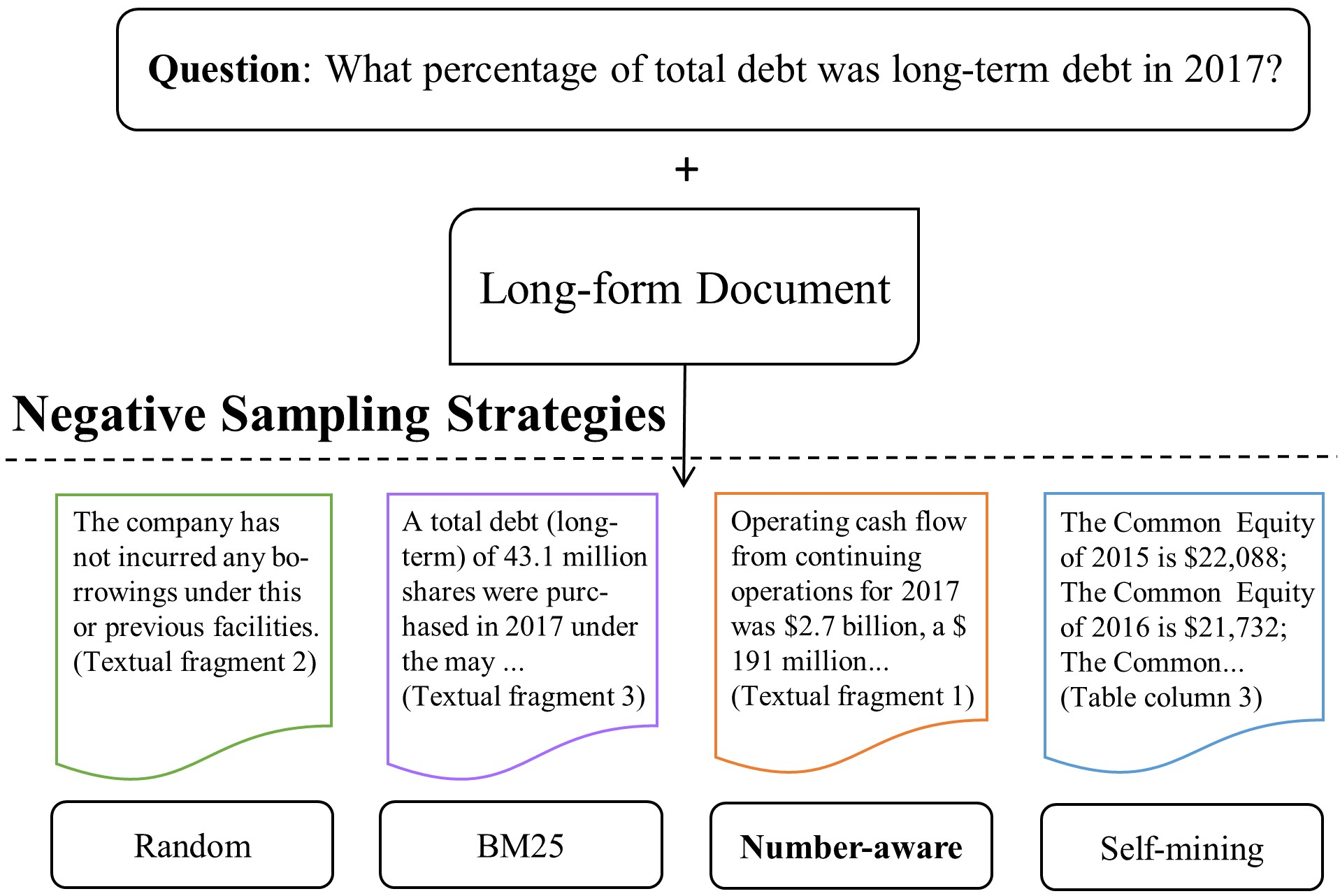}}
    \caption{The novel hard negative sampling strategy in APOLLO. The facts for four methods are sampled from the same document in Figure \ref{figure:example}. We compared our sampling method with three frequently conventional methods: Random, BM25 and Self-mining.}
    \label{figure:retriever}
\end{figure}

\section{Methodology}
In this section, we describe our approach in detail. Similar to \citet{chen2021finqa}, we utilize the retriever-generator framework and train each module independently. However, our approach differs from previous work \cite{zhang2022robustly, zhang2022elastic} in two ways: (1) the retriever is trained using number-aware negative sampling, and (2) the generator is trained with target program augmentation and subsequently refined using consistency-based reinforcement learning.

\subsection{Task Definition}
\label{sec:task_definition}

Given a question $q$ and a long-form document $d$ consisting of textual and structured-table facts, long-form numerical reasoning aims to generate a program that can be executed to get the right answer. Since the supporting document is too long to handle, this task can be decomposed into two sub-tasks: 1) retrieving the key facts from document $d$; 2) generating the program based on retrieved facts and $q$. The objective can be written as:
\begin{equation}
\begin{aligned}
P(G|q;d) = P(G|q,F;d)P(F|q;d)  
\end{aligned}
\end{equation}
where $G = [w_0,w_1,...,w_l]$ donate the golden program sequence consisting of $l$ program tokens $w$, $F=\{f_1,f_2,...,f_m\}$ donates the key fact set where each fact $f_i$ potentially contribute to answer the question $q$. 
In a typical retriever-generator framework~\cite{chen2021finqa}, the retriever learns to maximize $P(F|q;d)$, and the generator aims to maximize $P(G|q;d)$. 





\subsection{Retriever}
\label{sec:retriever}

The retriever is based on the sequence-pair classification model \citet{nogueira2019passage}. The question $q$ and each fact $f_i$ are concatenated to a BERT \cite{bert} transformer. A pooling layer followed by a linear layer is adopted to obtain the score $s^m$ for each fact:
\begin{equation}
\begin{aligned}
s^m &= W_l^{T}cls(Encoder(concat(q,f_i)))
\end{aligned}
\end{equation}
where $f_i \in F$ and $cls()$ extracts BERT’s hidden vector at token $[CLS]$, $W_l$ is a projection vector.
\paragraph{Number-aware Negative Sampling}
The score $s^m$ is used to rank each fact. To compute the $s^m$ accurately, the retriever is trained on both positive and negative facts. Previous works \cite{chen2021finqa,convfinqa,wang2022numerical} randomly sample negative facts in the given long-form document. However, since numerical reasoning is directly related to numbers, intuitively, facts without numbers are less contributing. As shown in Figure~\ref{figure:example}, the two green-highlighted facts are both numerical facts, which are the primary source of parameters in the gold program. In contrast, the red-highlighted non-numerical fact contributes less to the gold program and answer. We expect the retriever to focus more on numerical facts. 

In order to effectively train the retriever, we utilize a number-aware negative sampling strategy which involves extracting numerical facts from negative examples within long-form documents and randomly selecting a subset of these numerical facts.

\begin{figure*}[t]
\centerline{\includegraphics[width=15.5cm,keepaspectratio]{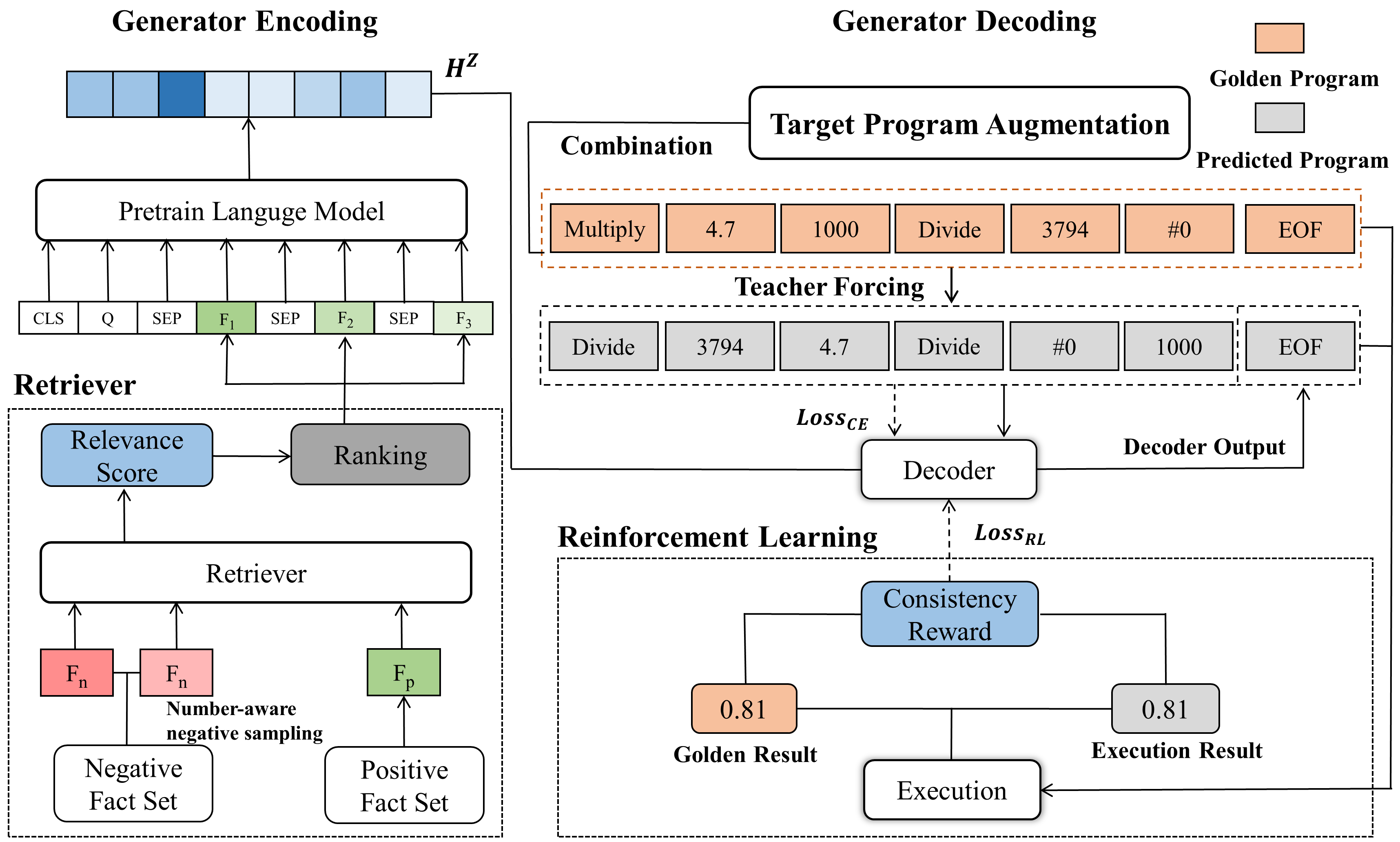}}
    \caption{The overall architecture of retriever-generator framework with APOLLO. $F_n$ and $F_p$ denotes negative facts and positive facts for training, respectively, and $F_1,F_2,F_3$ represent the retrieved facts. We use golden program in Figure~\ref{figure:example} as an example. The left portion of the figure illustrates the retriever and encoding process for the generator, while the right portion illustrates the complete process of generating the "EOF" token, implementing target program augmentation, and consistency-based reinforcement learning. The generator utilizes cross-entropy to supervise the generation of predicted programs, using both the golden program and programs generated through target program augmentation as reference. Then, APOLLO samples consistent program and executes with golden program to obtain the execution and golden results, which are then used in Equation~\ref{equation:rl} to calculate the consistent reward. This consistent reward is then employed to update all parameters.}
    \label{figure:generator}
\end{figure*}

\begin{figure}[t]
\centerline{\includegraphics[width=7cm,keepaspectratio]{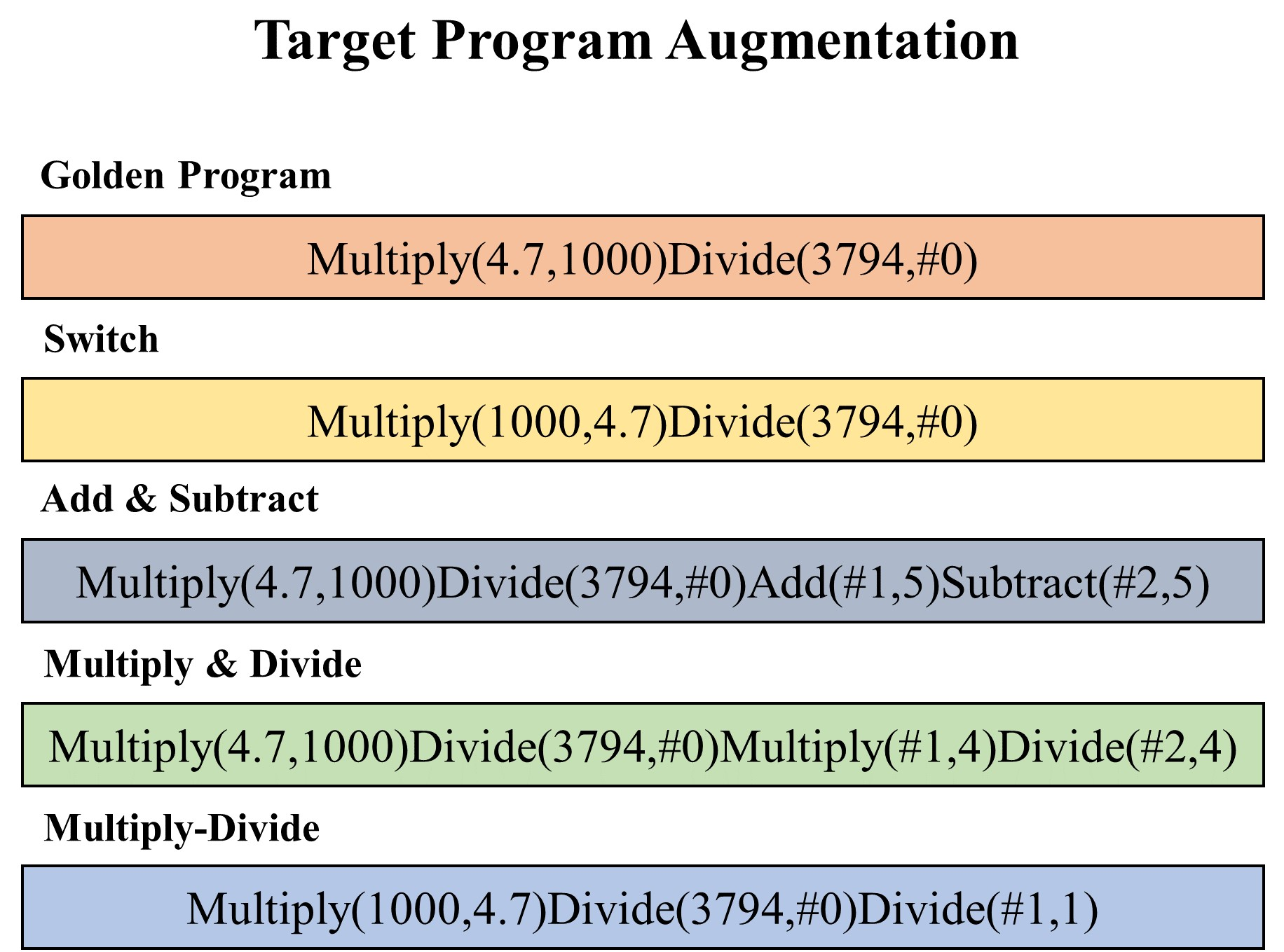}}
    \caption{The specific form of four target program augmentation construction. All the results of the programs generated by target program augmentation are the same as the golden program.}
    \label{figure:TPA}
\end{figure}

Figure~\ref{figure:retriever} illustrates the differences between number-aware negative sampling and conventional negative sampling strategies. These conventional strategies include (1) Random: selecting a random fact from the corpus, (2) BM25: selecting the most similar fact returned by BM25~\cite{bm25} algorithm, and (3) Self-mining: selecting the highest ranking fact retrieved by a well-trained retriever. It can be observed that random sampling does not guarantee sufficient high-quality negative facts. Although BM25 constructs negative facts that have a high overlap with the characters in the question, they tend to contain few numbers and are less supportive of generating programs. Furthermore, the self-mining strategy is prone to sampling duplicate negative facts (e.g., from the same table column), making the negative facts less informative in discriminating key facts.




\subsection{Generator}
\label{sec:generator}

The typical generator consists of a pre-trained encoder and a decoder. First, the question and Top-k retrieved facts are concatenated together to a BERT encoder to produce token-level contextualized embeddings:
\begin{equation}
H^{c} = Encoder(concat(q,Z)
\end{equation}
where $Z = \{f_1,f_2,...,f_k\}$
 is the Top-K retrieved facts set. Then, the decoder takes the embedding $H^{z}$ as input and decodes the numerical reasoning program step by step:
\begin{equation}
P(w_t|q) = Decoder(w_{t-1},H^{c})
\end{equation}
where $w_t$ is generated token in step $t$. 

Cross-entropy loss is adopted to supervise the generator. In cross-entropy, the expected value at each position is the corresponding token from the ground-truth program. However, due to the program consistency issue, the supervision is sometimes misleading. For example, in Figure~\ref{figure:example}, the ground truth is \texttt{Multiply(4.7,const\_1000)}, \texttt{Divide(3794,\#0)}. Then, although \texttt{Divide(3794, 4.7)}, \texttt{Divide(\#0,const\_1000)} also obtains the correct execution result, since it does not have an exact string match, it will be wrongly penalized using cross entropy loss.

To accurately compute $P(G|q;d)$ for all consistent programs, we first expand the ground truth set by target program augmentation. However, there are an infinite amount of programs for calculating a correct answer, this paper focuses on regular construction forms for simplicity. Additionally, we also utilize consistency-based reinforcement learning to directly optimize for the expected correctness of the execution result. 

\paragraph{Target Program Augmentation}
\label{sec:TPA}
In generating the program, we have ten program computation operators, including \texttt{Add}, \texttt{Subtract}, \texttt{Multiply}, \texttt{Divide}, and so on. (Other operators and their specific parameters and outputs can be found in Appendix \ref{appendix:operator_definition}). We adopt four target program augmentation methods: Switch, Add \& Subtract, Multiply \& Divide, and Multiply-Divide. These four methods details are the following:

\textbf{Switch} For the  \texttt{Add} and \texttt{Multiply} operators, the two arguments are interchangeable. Heuristically, for programs with these two operators in the ground truth, we can swap these two parameters to increase the amount of data. Moreover, a program that has $n$ operators with addition or multiplication operators can create $2^{n}-1$ more new samples for training.

\textbf{Add \& Subtract} For the same program, the result does not change after adding and subtracting a random constant at the end. 

\textbf{Multiply \& Divide} Similar to the previous construction, the result does not change after multiplying and dividing a random constant at the end. 

\textbf{Multiply-Divide} The same program multiplies or divides a constant one does not change its result.

We use golden program in Figure \ref{figure:example} as an example to illustrate the specific format of the four target program augmentation demonstrated in Figure \ref{figure:TPA}.



\paragraph{Consistency-based Reinforcement Learning}
\label{sec:rl}
Instead of using teacher forcing at each step of program generation, the next token is obtained by sampling from the output distribution. At the end of the generation procedure, the generated program is executed and compared against the correct answer to determine a reward. Let $G_g$ denotes the program generated by the generator and $G_T$ denotes the ground truth program corresponding to the question. We define the reward $R(G_g, G_T)$ as
\begin{equation}
R\left(G_g, G_{T}\right)=\left\{\begin{array}{ll}
-2, & \text{U.E.P} \\
-1, & \text{E.P but wrong answer} \\
+1, & \text{E.P and right answer} \\
\end{array}\right.
\label{equ:reward}
\end{equation}
where $U.E.P$ stands for unexecutable program and $E.P$ stands for executable program.
The reinforcement learning loss is the negative expected reward over possible generated program $L^{RL} = -\mathbb{E}_{w}[R(G_g, G_{T})]$. Inspired by \citet{zhong2017seq2sql}, we derive the policy gradient for $L^{RL}$:
\begin{equation}
\label{equation:rl}
\begin{aligned}
\nabla L_{\Theta}^{\text {RL }} &=-\nabla_{\Theta}\left(\mathbb{E}_{w \sim p_{w}}\left[R\left(G_g, G_{T}\right)\right]\right) \\
& \approx-R\left(G_g, G_{T}\right) \nabla_{\Theta} \sum_{t}\left(\log p_{w}\left(w_{t} ; \Theta\right)\right)
\end{aligned}
\end{equation}
where $p_{w}$ denotes the probability of choosing token $w_t$ during decoding time step $t$. Moreover, we approximate the expected gradient using a single Monte-Carlo sample $w$.

The training process with consistency-based reinforcement learning is shown in Figure \ref{figure:generator}. While the golden program and the generated program differ at the character-level, they are identical at the consistency-level, leading to a positive reward. However, if only used cross-entropy training, the predicted program generated by decoding may be incorrectly penalized.

\begin{table*}[]
\resizebox{\linewidth}{!}{
\begin{tabular}{l|cc|cc|cc|cc}
\hline
\multirow{2}{*}{\textbf{Model}} & \multicolumn{2}{c|}{\textbf{FinQA(dev)}} & \multicolumn{2}{c|}{\textbf{FinQA(test)}} & \multicolumn{2}{c|}{\textbf{ConvFinQA(dev)}} & \multicolumn{2}{c}{\textbf{ConvFinQA(test)}} \\
 & \textbf{Exe Acc} & \textbf{Prog Acc} & \textbf{Exe Acc} & \textbf{Prog Acc} & \textbf{Exe Acc} & \textbf{Prog Acc} & \textbf{Exe Acc} & \textbf{Prog Acc} \\ 
\hline
\multicolumn{9}{c}{\textit{Prompting}} \\ 
\hline
BloombergGPT \cite{bloombergGPT} & - & - & - & - & 43.41 & - & - & - \\ 
GPT-3.5-turbo \cite{chatgpt} & - & - & 48.56 & - & 59.86 & - & - & - \\
GPT-4 \cite{gpt4} & - & - & 68.79  & - & 76.48 & - & - & - \\ 
Program-of-Thought \cite{pot} & - & - & 68.10 & - & 67.30 & - & - & - \\
\hline
\multicolumn{9}{c}{\textit{Fine-tuning}} \\ 
\hline
GPT-2~\cite{gpt} & - & - & - & - & 59.12 & 57.52 & 58.19 & 57.00 \\
T-5~\cite{t5} & - & - & - & - & 58.38 & 56.71 & 58.66 & 57.05 \\ 
Retriever+NeRd~\cite{nerd} & 47.53 & 45.37 & 48.57 & 46.76 & - & - & - & - \\
Longformer~\cite{beltagy2020longformer} & 23.83 & 22.56 & 21.90 & 20.48 & - & - & - & - \\
FinQANet~\cite{chen2021finqa} & 61.22 & 58.05 & 61.24 & 58.86 & 68.32 & 67.87 & 68.90 & 68.24 \\
ELASTIC~\cite{zhang2022elastic} & 65.00 & 61.00 & 62.16 & 57.54 & - & - & - & - \\
DyRRen~\cite{dyrren} & 66.82 & 63.87 & 63.30 & 61.29 & - & - & - & - \\
TabT5*~\cite{tablet5} & - & - & 70.79 & 68.00 & - & - & - & - \\
CellRetriever+UniLM*~\cite{wang2022numerical} & - & - & 68.00 & 65.21 & - & - & - & -\\ 
\hline
APOLLO & 69.70 & 65.91 & 67.99 & 65.60 & 76.47 & 74.14 & 76.00 & 74.56 \\
- Ensemble model & \textbf{72.91} & \textbf{70.83} & \textbf{71.07} & \textbf{68.94} & \textbf{78.46} & \textbf{75.91} & \textbf{78.76} & \textbf{77.19} \\ 
\hline
General Crowd Performance & - & - & 50.68 & 48.17 & - & - & 46.90 & 45.52 \\
Human Expert Performance & - & - & 91.16 & 87.49 & - & - & 89.44 & 86.34 \\ 
\hline
\end{tabular}
}
\caption{Performance comparisons on the dev set, test set of FinQA and dev set, private test set of ConvFinQA. The pre-trained models utilized in rows 5 to 11 of the table are all RoBERTa-large, except for TabT5 and CellRetriever+UniLM, which use T5 and UniLM~\cite{dong2019unified} respectively. The missing data in the table is due to the fact that many works do not report their results. * denotes the results of ensemble models, since many works only report their ensemble model results.}
\label{table:generator}
\end{table*}

\section{Experiment}
In this section, we mainly introduce the datasets we evaluate in our long-form numerical reasoning task and the experimental effect of our work to show the advantages of our methods from an experimental comparison view. 

\subsection{Datasets}
We conduct evaluation experiments on two datasets: FinQA \cite{chen2021finqa} and ConvFinQA \cite{convfinqa}.

\paragraph{FinQA}
FinQA is a dataset of numerical reasoning over long-form financial data, containing $8,281$ financial reports, along with their QA pairs and annotated numerical reasoning processes by eleven finance professionals based on the earnings reports of S\&p $500$ companies \cite{zheng2021global}. The data is released as training ($6,251$), dev ($883$), and test ($1,147$) following a 75\%/10\%/15\% split. The long-form financial documents contained heterogeneous data (structured and unstructured data such as tables and texts) and compounded many financial terms in questions, e.g., "Shares vested", and "Pre-tax earnings" which is challenging in this long-form numerical reasoning task.

\paragraph{ConvFinQA}
ConvFinQA (Conversational Finance Question Answering) is a dataset of conversational long-form numerical reasoning over financial data, containing $3,892$ conversations consisting of $14,115$ questions and annotated by expert annotators to compose the question based on the simulated conversing flow. The data has split into $3,037$/$421$/$434$ for train/dev/test sets. However, the reasoning chains throughout the conversation pose great challenges for the models to learn when to refer to or discard the conversation history and how to assemble the reasoning path.

\begin{table*}[t]
\centering
\begin{tabular}{lc|cc|cc|cc}
\hline
\multirow{2}{*}{\textbf{Model}} & \multirow{2}{*}{\textbf{Pre-training}} & \multicolumn{2}{c|}{\textbf{FinQA(dev)}} & \multicolumn{2}{c|}{\textbf{FinQA(test)}} & \multicolumn{2}{c}{\textbf{ConvFinQA(dev)}} \\
                       &                                 & \textbf{R@3}  & \textbf{R@5}                     & \textbf{R@3}   & \textbf{R@5}                     & \textbf{R@3}              & \textbf{R@5}             \\ \hline
\multicolumn{1}{l}{FinQANet~\cite{chen2021finqa} } & \multicolumn{1}{c|}{RoBERTa}          & 91.30    & \multicolumn{1}{l|}{93.89}  & 89.82    & \multicolumn{1}{l|}{93.22}  & 88.95                & 92.74              \\
\multicolumn{1}{l}{FinQANet~\cite{chen2021finqa}}  & \multicolumn{1}{c|}{DeBERTa-v3}           &  92.03    & \multicolumn{1}{l|}{95.06}   &   90.37    & \multicolumn{1}{l|}{93.78}   &      89.49            &    92.91             \\
\multicolumn{1}{l}{Ant Risk AI~\cite{zhang2022robustly} }  & \multicolumn{1}{c|}{RoBERTa}           & 91.54     & \multicolumn{1}{l|}{95.11}   &   90.16    & \multicolumn{1}{l|}{94.12}   &      -            &      -           \\
\multicolumn{1}{l}{- Ensemble model}  & \multicolumn{1}{c|}{RoBERTa}           &   92.63   & \multicolumn{1}{l|}{95.89}   &  90.77     & \multicolumn{1}{l|}{94.33}   &          -        &        -         \\ \hline
\multicolumn{1}{l}{APOLLO}  & \multicolumn{1}{c|}{RoBERTa}           &  93.58    & \multicolumn{1}{l|}{95.62}   &    91.76   & \multicolumn{1}{l|}{93.95}   &          91.67        &       94.56          \\
\multicolumn{1}{l}{APOLLO}  & \multicolumn{1}{c|}{DeBERTa-v3}           &  94.22    & \multicolumn{1}{l|}{96.08}   &   92.37    & \multicolumn{1}{l|}{94.49}   &      92.18            &      95.01           \\
\multicolumn{1}{l}{- Ensemble model}  & \multicolumn{1}{c|}{DeBERTa-v3}           &  \textbf{95.03}    & \multicolumn{1}{l|}{\textbf{96.54}}   &   \textbf{93.31}   & \multicolumn{1}{l|}{\textbf{94.98}}   &    \textbf{92.40}              &      \textbf{95.15}           \\ \hline
\end{tabular}
\caption{The experimental results of retriever Recall Top-3 and Top-5 on the dev set and test set in FinQA and only dev set on ConvFinQA. ConvFinQA only has private test set available currently which dos not have ground truth for retrieved facts. All pre-training models are large-size models.}
\label{table:retriever}
\end{table*}

\subsection{Baselines}
We compare our model with several competitive models based on prompting and fine-tuning. \textbf{Prompting-based} methods: (1) \textbf{BloombergGPT} \cite{bloombergGPT}, which is a 50-billion parameter LLM that was purpose-built from scratch for finance. (2) \textbf{GPT-3.5-turbo} \cite{chatgpt}, which is a 175B parameter LLM trained on diverse textual data. (3) \textbf{GPT-4} \cite{gpt4}, which is a large-scale multimodal model capable of accepting image and text inputs and producing text output. (4) \textbf{Program-of-Thought} \cite{pot}, which uses Codex \cite{codex} to generate text and programming language statements, and finally an answer. The GPT-3.5-turbo and GPT-4 settings are zero-shot, and the results are directly from \citep{chatgpt_few_shot}.

\textbf{Fine-tuning-based} methods: (1) \textbf{FinQANet} \cite{chen2021finqa}, which utilizes a retriever-generator framework to generate programs based on retrieved facts. (2) \textbf{NeRd} \cite{nerd}, which employs a BERT-based pointer-generator model to generate symbolic nested programs. (3) \textbf{Longformer} \cite{beltagy2020longformer}, which inputs the entire contents of long-form documents and generates programs. (4) \textbf{GPT-2} \cite{gpt}, which uses the GPT-2 model with prompts to generate programs. (5) \textbf{T5} \cite{t5}, which is similar to GPT-2 and utilizes the T5 model with prompts to generate programs. (6) \textbf{CellRetriever+UniLM} \cite{wang2022numerical}, which employs both cell and row retrievers to retrieve facts and integrates multiple generators to generate programs. (7) \textbf{ELASTIC} \footnote{We find a serious data leak and we re-do their experiment based on their released code. The data leak results of their model and APOLLO are shown in Appendix \ref{appendix:data_leakage}.} \cite{zhang2022elastic}, which utilizes an adaptive symbolic compiler to generate programs. (8) \textbf{Ant Risk AI} \cite{zhang2022robustly}, which develops models with different specialized capabilities and fuses their strengths to retrieve and generate programs. (9) \textbf{TabT5} \cite{tablet5}, which uses the T5 model pre-trained on Wikipedia tables to generate programs. (10) \textbf{DyRRen} \cite{dyrren}, which enhances each generation step through a dynamic reranking of retrieved facts using a retriever-reranker-generator framework. (11) \textbf{Human performance}, which includes both experts and non-experts in the FinQA and ConvFinQA datasets. The results are taken from the original paper \cite{chen2021finqa, convfinqa}.

\begin{table}[t]
\resizebox{\linewidth}{!}{
\begin{tabular}{lc|ll|ll}
\hline
\multirow{2}{*}{\textbf{Type}} & \multirow{2}{*}{\textbf{\#N}} & \multicolumn{2}{c|}{\textbf{FinQA}} & \multicolumn{2}{c}{\textbf{ConvFinQA}} \\ 
            &             & \textbf{R@3}          & \textbf{R@5}         & \textbf{R@3}            & \textbf{R@5}           \\ \hline
Random                & 3                         & 90.37            & 94.07        & 89.49              & 92.91             \\
BM25                  & 3                         & 88.20            & 92.32        & 88.21              & 91.40             \\
Number-aware          & 3                         & 92.02            & 94.19        & \textbf{92.18}       & \textbf{95.01}    \\ \hline
Random                & 2                         & 89.56            & 93.63           & 89.27              & 92.63             \\
Random                & 4                         & 89.87            & 93.74           & 89.15              & 92.80             \\
Random                & 5                         & 89.75           & 93.67         & 88.91            & 92.78             \\ \hline
BM25                  & 2                           & 88.86            & 92.81        & 86.02              & 88.12              \\
BM25                  & 4                           & 88.35            & 91.29        & 87.57              & 89.11               \\
BM25                  & 5                          & 88.83            & 92.36        & 86.37              & 89.95               \\ \hline
Self-mining  \& R        & 3                          & 89.47            & 94.01        & 88.62              & 90.82             \\
Self-mining  \& B        & 3                          & 84.28            & 87.24        & 86.91              & 88.08      \\
Self-mining   \& N       & 3                        & 91.25            & 93.73        & 88.31              & 90.27             \\ \hline
Number-aware          & 2                         & 91.98            & 94.04           & 91.87              & 94.68             \\
Number-aware          & 4                         & \textbf{92.37}       & \textbf{94.49}       & 91.92      & 94.70     \\
Number-aware          & 5                         & 92.01          & 94.23           & 91.91          & 94.26             \\ \hline
\end{tabular}
}
\caption{The results of different negative sampling strategies on test set in FinQA and dev set in ConvFinQA. Self-mining  \& R, B, and N denote using Self-mining with Random, BM25, and Number-aware negative sampling, respectively. \#N: Ratio of positive and negative samples for training.}
\label{table:number-aware}
\end{table}

\begin{table}[htbp]
\resizebox{\linewidth}{!}{
\begin{tabular}{l|cc|cc}
\hline
\multirow{2}{*}{\textbf{Model}} & \multicolumn{2}{c|}{\textbf{FinQA}} & \multicolumn{2}{c}{\textbf{ConvFinQA}} \\ 
                       & \textbf{Exe Acc}     & \textbf{Prog Acc}     & \textbf{Exe Acc}       & \textbf{Prog Acc}      \\ \hline
Generator              & 66.95       & 64.62        & 75.64         & 73.13         \\
w. Switch                & 67.07       & 64.62            & 75.78             & 73.13             \\
w. Add \& Sub       & 67.25      &  64.95            &  75.82              &  73.54             \\
w. Mul \& Divide    & 67.46       &  65.07           &    75.95            &  73.70             \\
w. Mul-Divide       & 66.95       &   64.62           &    75.64            &   73.13            \\ \hline
w. RL    & 67.36       &  65.14           &    76.14            &  73.61             \\ 
w. RL \& TPA   & \textbf{67.99}       &  \textbf{65.60}           &    \textbf{76.47}            &  \textbf{74.14}             \\ \hline
\end{tabular}
}
\caption{
The performances of APOLLO with consistency-based reinforcement learning and different target program augmentation methods. RL \& TPA denotes combine two approches, which performs best on both datasets.}
\label{table:RL_TPA}
\end{table}

\subsection{Evaluation Metrics}
 \paragraph{Retriever}
In retriever, we use Recall Top-3 and Recall Top-5 to evaluate our model. This metric evaluates the retrieval result by determining the percentage of correct positive predictions out of all positive predictions. However, since there may be more than one positive prediction in each sample, we assume that the first $N$ predictions in the recall top-N are all positive predictions.
\paragraph{Generator}
In generator, we use Execution Accuracy and Program Accuracy to evaluate our model. Execution Accuracy evaluates the model by calculating the accuracy between the predicted program result and the golden executable result. Program Accuracy calculates the accuracy of the operators and operands between the predicted program and the golden program.

\subsection{Implementation Details}
Our model is implemented using Pytorch \cite{paszke2019pytorch} and Transformer \cite{wolf2020transformers}, and then trained on a server with two NVIDIA Tesla A100 GPUs of 40G memory. For retriever, we use RoBERTa-large and DeBERTa-v3-large as the classifier. We take the top-3 ranked facts as the retriever results. Training epochs are set to 50 and batch size for all datasets is 8. The initial learning rate is set to 9e-6, and we schedule the learning rate to warm up at the beginning and gradually decrease the learning rate during training. For generator, we adopt RoBERTa as the encoder, the initial learning rate is set to 1e-5 and then adopt learning rate scheduler. For consistency-based reinforcement learning, we adopt continual learning to continue train based on our best performance model. For all models, the maximum sequence length is set to 512. Besides, we use Adam as optimizer \cite{kingma2014adam} to update the parameters of the models and clip the gradient every iteration to prevent gradient explosion as well as applying weight decay to prevent over-fitting.

\begin{figure}[]
    \centerline{\includegraphics[width=8cm,keepaspectratio]{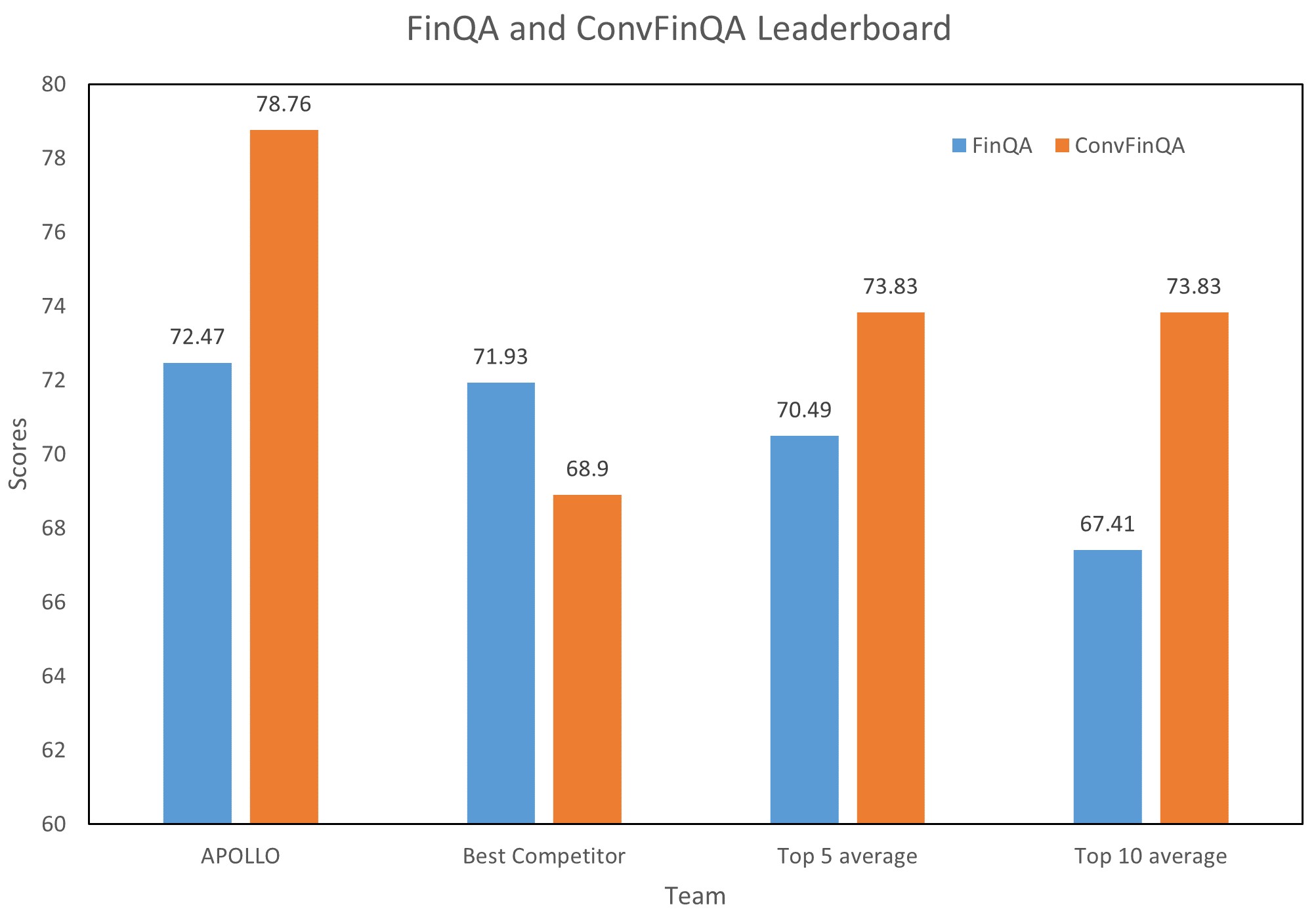}}
    \caption{Performance comparisons on the private test set of the FinQA and ConvFinQA. We report APOLLO, best competitor, Top 5 average and Top 10 average scores on both leaderboard. At the time of submission (25 Nov. 2022), APOLLO has achieved state-of-the-art in both leaderboards. }
    \label{figure:private}
\end{figure}

\subsection{Main Results}

Table \ref{table:generator} presents the generator performance of APOLLO and baselines on FinQA and ConvFinQA. APOLLO achieves the highest scores on both datasets, with 71.07 execution accuracy, 68.94 program accuracy on FinQA test set and 78.46 points execution accuracy, 75.91 program accuracy on ConvFinQA dev set. 

Table \ref{table:retriever} presents the retriever performances of APOLLO and baselines on FinQA and ConvFinQA. Overall, APOLLO achieves the highest scores on both datasets. A comparison of APOLLO to the baseline FinQANet demonstrates a notable advantage, with 2.00 points higher Recall 3 and 0.71 Recall 5 on FinQA, and 2.69 points higher Recall 3, 2.1 Recall 5 on ConvFinQA. However, some works \cite{wang2022novel,wang2022numerical, dyrren} do not report their retriever performance on dev or test sets and do not release their source code, so they are not included in the retriever results table. Additionally, our single model has achieved a significant improvement over previously published ensemble models on both the dev and test sets.

In general, APOLLO achieves the best performance on both FinQA and ConvFinQA leaderboard of the private test set, shown in Figure \ref{figure:private}.


\subsection{Ablation Study}
APOLLO comprises three crucial components: number-aware negative sampling in retriever training; consistency-based reinforcement learning and target program augmentation in generator training. To gain a deeper understanding of our model, we conduct an extensive ablation study to investigate the impact of these three modules individually.

\paragraph{Results of Number-aware Hard Negative Training}
Table~\ref{table:number-aware} illustrates the performance of different methods for constructing negative facts. The results indicate that our number-aware negative sampling strategy surpasses random, BM25 and self-mining on both long-form numerical reasoning datasets. Furthermore, we experimentally investigate the impact of the ratio of positive to negative facts on training. We conduct experiments with positive to negative fact ratios of 3, 4 and 5, and find that the ratio has a substantial influence on the results. It demonstrates that the positive and negative fact ratio has a great influence on training. The best results for the random method, BM25 and self-mining are obtained with ratio of 3, where the ratio changes, the performance decreases.

\paragraph{Results of Consistency-based Reinforcement Learning Training and Target Program Augmentation}
Table \ref{table:RL_TPA} presents the results of utilizing consistency-based reinforcement learning and target program augmentation in generator training. Noteworthy, among the four target program augmentation methods, the Multiply \& Divide methods demonstrated the greatest enhancement on both datasets. We suspect this is due to the operation distributions, where \texttt{Multiply}, and \texttt{Divide} have the total distributions of 51.11\%. These two operations contribute largely to dataset, and there is also a high percentage of simultaneous between multiplication and division in the program, leading to more stable model convergence. However, the target program augmentation method in ConvFinQA receives poor performance compared with FinQA, we suspect that many answers in ConvFinQA are directly obtained from the original long-form document, making the target program augmentation ineffective. Additionally, the consistency-based reinforcement learning method also yielded significant performance gains on both datasets. Furthermore, the best results are obtained by combining the two approaches, with target program augmentation followed by reinforcement learning training.

Moreover, APOLLO increases the diversity of the generated programs, shown in Appendix \ref{appendix:diversity}.

\section{Conclusion}
This work presents an optimized training approach for long-form numerical reasoning, APOLLO. Unlike previous work, APOLLO enhances the retriever with number-aware negative sampling strategy to better classify numerical facts and the generator with consistency-based reinforcement learning as well as target program augmentation to generate more accurate and diverse programs. As a result, APOLLO achieves the state-of-the-art results on both FinQA and ConvFinQA leaderboards, which significantly outperforms all the other models.

\section*{Limitations}
APOLLO mainly has two limitations:
\begin{itemize}
\item The negative sampling strategy in our retriever training, which is sensitive to numerical data, is appropriate for tasks that involve a substantial amount of numerical information, particularly in financial domain. However, it is not suitable for tasks that have lower numerical content, as it lacks scalability in those cases.
\item The target program augmentation construction form is relatively basic. We will explore more advanced forms of construction in future work.

\end{itemize}

\bibliography{custom}
\bibliographystyle{acl_natbib}

\newpage

\appendix

\section{Appendix: Operator Definition}
\label{appendix:operator_definition}
Following by \citet{chen2021finqa}, we define all the operations in Table~\ref{table:operator}, which illustrates the specific operators and their operands.
The first four operators account for the largest distribution (about 94.29\%) of the entire dataset. The operation division has the highest frequency, as calculating ratios are common in financial analysis. Moreover, $none$ is a placeholder indicating that does not input another operand.

\section{Appendix: Data leakage}
\label{appendix:data_leakage}
Before May 11\footnote{\url{https://github.com/czyssrs/FinQA/commits/main/dataset}}, FinQA dataset suffered a serious data leakage. This is due to the fact that the ground-truth is directly utilized in the construction of the test data, resulting in an inflated recall 3 accuracy of the retriever (about 95\%) and ridiculously high execution accuracy of the generator. However, when testing on the private dataset of FinQA leaderboard, which doesn't have ground truth, the model performance is very poor. ELASTIC~\cite{zhang2022elastic} uses this wrong data, so we reported FinQANet, ELASTIC, and APOLLO based on the data leakage, as shown in Table~\ref{table:data_leakage}.

\begin{table}[h]
\centering
\begin{tabular}{l|c|c}
\hline
Model          & Exe Acc & Prog Acc \\ \hline
FinQANet       & 65.05   & 63.52    \\
ELASTIC        & 68.96   & 65.21    \\
APOLLO           & \textbf{72.44}   & \textbf{69.48}    \\ \hline
Gold-Retriever & 70.00   & 68.76    \\ \hline
\end{tabular}
\caption{The experimental results on test set in FinQA with data leakage. All the model are fine-tuning on RoBERTa-large. Gold-Retriever denotes the retrieved facts directly using ground-truth.}
\label{table:data_leakage}
\end{table}

\section{Appendix: Examples of Target Program Augmentation}
\label{appendix:example_TPA}
In Section~\ref{sec:generator}, we give a few examples of specific construction forms of target program augmentation, and to demonstrate that our method can generate programs more diversity.

\paragraph{Specific Constructed Form} 
We take a slightly more complex program as an example to explain exactly how we build target program augmentation. The four target program augmentation constructed forms are illustrated in Table \ref{table:construction_TPA}.

\paragraph{Diversity of Program Generation}
\label{appendix:diversity}
Most of the models trained by cross-entropy which can only generate programs with fixed templates, but since both FinQA and ConvFinQA are from real finance domain, our target program augmentation and reinforcement learning training is necessary for generate programs with more flexible. We trained several models with different hyper-parameters using our method and compared their outputs to ground-truth. The results are shown in Table~\ref{table:diversity}, which indicates that our model doesn't rely on templates, but generates more diverse and consistent identical programs.

\begin{table*}[t]
\centering
\begin{tabular}{l|c|c|l}
\hline
Name          & Operands                      & Output & Description                         \\ \hline
Add           & (number1,number2)             & number & add two numbers                     \\
Subtract      & (number1,number2)             & number & substract two numbers               \\
Multiply      & (number1,number2)             & number & multiply two numbers                \\
Divide        & (number1,number2)             & number & divide two numbers                  \\
Exp           & (number1,number2)             & number & calculate exponent:                 \\
Greater       & (number1,number2)             & bool   & return number1\textgreater{}number2 \\
Table-sum     & (table-header,none)           & number & the summation of one table row      \\
Table-average & (table-header,none)           & number & the average of one table row        \\
Table-max     & (table-header,none)           & number & the maximum number of one table row \\
Table-min     & (table-header,none)           & number & the minimum number of one table row \\ \hline
\end{tabular}
\caption{The operators and operands defined by \cite{chen2021finqa}}
\label{table:operator}
\end{table*}

\begin{table*}[t]
\centering
\resizebox{\linewidth}{!}{
\begin{tabular}{l|l}
\hline
Methods                 &  Programs          \\ \hline
Original                &  Add(101, 96), Multiply(Const\_10, 105), Add(\#0, \#1), Divide(\#2, Const\_3)                 \\ \hline
\multirow{7}{*}{Switch} &  Add(96, 101), Multiply(Const\_10, 105), Add(\#0, \#1), Divide(\#2, Const\_3)                \\
                        &  Add(101, 96), Multiply(105, Const\_10), Add(\#0, \#1), Divide(\#2, Const\_3)                \\
                        &  Add(96, 101), Multiply(105, Const\_10), Add(\#0, \#1), Divide(\#2, Const\_3)                 \\
                        &  Add(101, 96), Multiply(Const\_10, 105), Add(\#1, \#0), Divide(\#2, Const\_3)                 \\
                        &  Add(96, 101), Multiply(Const\_10, 105), Add(\#1, \#0), Divide(\#2, Const\_3)                 \\
                        &  Add(101, 96), Multiply(105, Const\_10), Add(\#1, \#0), Divide(\#2, Const\_3)                  \\
                        &  Add(96, 101), Multiply(105, Const\_10), Add(\#1, \#0), Divide(\#2, Const\_3)                 \\ \hline
Add \& Subtract         &  Add(101, 96), Multiply(Const\_10, 105), Add(\#0, \#1), Divide(\#2, Const\_3), Add(\#3, Const\_7), Subtract(\#4, Const\_7)           \\ \hline
Multiply \& Divide      &  Add(101, 96), Multiply(Const\_10, 105), Add(\#0, \#1), Divide(\#2, Const\_3), Multiply(\#3, Const\_4), Divide(\#4, Const\_4)        \\ \hline
Multiply-Divide         &  Add(101, 96), Multiply(Const\_10, 105), Add(\#0, \#1), Divide(\#2, Const\_3), Multiply(\#3, Const\_1)          \\ \hline
\end{tabular}
}
\caption{The specific form of four Target program augmentation construction.}
\label{table:construction_TPA}
\end{table*}

\begin{table*}[t]
\centering
\resizebox{\linewidth}{!}{
\begin{tabular}{l}
\hline
Programs                                                                       \\ \hline
Multiply(4.7, 1000), Divide(3794, \#0) (Original) \\ \hline
Divide(3794, 4.7), Divide(\#0, 1000) (RL) \\
Multiply(1000, 4.7), Divide(3794, \#0) (Switch) \\
Multiply(4.7, 1000), Divide(3794, \#0), Add(\#1, Const\_6), Subtract(\#2, Const\_6) (Add \& Subtract)                              \\
Multiply(4.7, 1000), Divide(3794, \#0), Multiply(\#2, Const\_4), Divide(\#3, Const\_4) (Multiply \& Divide)                         \\
Multiply(4.7, 1000), Divide(3794, \#0), Multiply(\#1, Const\_1) (Multiply-Divide)                                                           \\ \hline
\end{tabular}
}
\caption{Experiment of the programs generated by our model after reinforcement learning and target program augmentation training. Our model can generate programs with more diversity.}
\label{table:diversity}
\end{table*}

\end{document}